\def\year{2020}\relax
\documentclass[letterpaper,table]{article} 
\usepackage{aaai20}  
\usepackage{times}  
\usepackage{helvet}  
\usepackage{courier}  
\usepackage[hyphens]{url}  
\usepackage{graphicx}  
\usepackage{amsfonts}
\usepackage{amsmath}
\usepackage{float}
\usepackage{multirow}
\usepackage{tabularx}
\usepackage{xcolor}
\usepackage{makecell}
\usepackage{amssymb} 
\usepackage{subfig}
\usepackage{booktabs}
\usepackage{algorithm}
\usepackage{algorithmic}
\usepackage{amsthm}
\DeclareMathOperator*{\argmax}{arg\,max}
\newcolumntype{x}[1]{>{\centering\arraybackslash\hspace{0pt}}p{#1}}
\newcommand{\matr}[1]{\mathbf{#1}}  
\newcommand{\vect}[1]{\mathbf{#1}}  

\begin{document}

\title{Question Guided Modular Routing Networks for Visual Question Answering}
\author{Yanze Wu\textnormal{\textsuperscript{1}},
Qiang Sun\textnormal{\textsuperscript{1}},
Jianqi Ma\textnormal{\textsuperscript{1}},
Bin Li\textnormal{\textsuperscript{1}},
Yanwei Fu\textnormal{\textsuperscript{1}},
Yao Peng\textnormal{\textsuperscript{2}},
Xiangyang Xue\textnormal{\textsuperscript{1}}\\
\textsuperscript{1}Fudan University,
\textsuperscript{2}Qiniu Inc\\
\{wuyanze123,sunqiang85\}@gmail.com, \{majq16,libin,yanweifu\}@fudan.edu.cn, pengyao@qiniu.com, xyxue@fudan.edu.cn\\
}

\maketitle

\begin{abstract}
This paper studies the task of Visual Question Answering (VQA), which is topical in Multimedia community recently. Particularly, we explore two critical research problems existed in VQA: (1) efficiently fusing the visual and textual modalities; (2) enabling  the visual reasoning ability of VQA models in answering complex questions.
To address these challenging problems, a novel  Question Guided Modular Routing Networks (QGMRN) has been proposed in this paper. Particularly, 
The QGMRN is composed of visual, textual and routing network. 
The visual and textual network serve as the backbones for the generic feature extractors of visual and textual modalities. QGMRN can fuse the visual and textual modalities at multiple semantic levels. 
Typically,  the visual reasoning is facilitated by the routing network in a discrete and stochastic way by using Gumbel-Softmax trick for module selection.
When the input reaches a certain modular layer, routing network newly proposed in this paper, dynamically selects a portion of modules from that layer to process the input depending on the question features generated by the textual network. 
It can also learn to reason by routing between the generic modules without additional supervision information or expert knowledge. Benefiting from the dynamic routing mechanism, QGMRN can outperform the previous classical VQA methods by a large margin and achieve the competitive results against the state-of-the-art methods. Furthermore, attention mechanism is integrated into our QGMRN model and thus can further boost the model performance. Empirically, extensive experiments on the CLEVR and CLEVR-Humans datasets validate the effectiveness of our proposed model, and the   state-of-the-art performance has been achieved.
\end{abstract}

\section{Introduction}
Visual Question Answering (VQA)~\cite{antol2015vqa} is a task that,
given an image and question pair, the model can reason and answer the question by visual information. It is a popular topic in recent years
that has connected the computer vision and natural language processing
(NLP). VQA faces two major challenges: 1) How to better
fuse the visual and textual modalities; 2) How to make the VQA model have the reasoning ability to answer more complex questions.

For challenge (1), in order to fuse the visual and textual modalities, the most common 
VQA paradigm is to first extract the visual features from a modern CNN (\textit{e.g.,} VGG~\cite{simonyan2014very} or ResNet~\cite{he2016deep}) and textual features from a RNN (\textit{e.g.,} LSTM~\cite{hochreiter1997long}
or GRU~\cite{cho2014learning}) separately,  and then fuse them into a common
latent space~\cite{kazemi2017show,anderson2018bottom,malinowski2015ask,Fukui2016MultimodalCB,wu2018object}.
Feature fusion is explored from simple
concatenation operation~\cite{zhou2015simple} to advanced joint embedding
techniques, such as MLB~\cite{Kim2016MultimodalRL}, MCB~\cite{Fukui2016MultimodalCB}
and MUTAN~\cite{ben2017mutan}, and attention mechanisms such as one-hop attention~\cite{kazemi2017show}, multi-hop attention~\cite{yang2016stacked} and bottom-up attention~\cite{anderson2018bottom}.
However, no matter how powerful these fusion methods are, 
they can only fuse two modalities at a high semantic level, rather than multiple semantic levels, because the fused features (\textit{i.e.,} the extracted visual and textual features) are at a high semantic level.

For challenge (2), to endow the VQA model with reasoning ability is difficult because although CNN (\textit{e.g.,} ResNet) is very powerful, it does not support reasoning natively.
To make the VQA model support reasoning, one important line of work~\cite{andreas2016neural,hu2017learning,johnson2017inferring,mascharka2018transparency,mao2019neuro} builds upon the modular networks. By leveraging the compositional structure of neural language questions, these methods 
first decompose the question into modular subquestions, 
then pick out the modular networks that are designed for solving these subquestions, 
and finally answer the question by executing these modular networks on the image according to the layout learned from the question. 
These explicit methods have strong interpretability as the layout learned from the question reflects the answering and reasoning process. However, analyzing the subquestion set and designing corresponding modules need expert knowledge, most of these methods even require extra supervision information during their training process.

In this paper, we show that even without expert knowledge and extra supervision information,
we can still make the model support visual reasoning. Specifically, we propose a novel model called Question Guided Modular Routing Networks (QGMRN) that consists of three sub-networks: \textit{visual network}, \textit{textual network} and \textit{routing network}. The visual network is based on a generic modular network that each layer of the network is composed of some generic modules, as the module granularity changes, different modular networks can be spawned. When the input reaches a certain layer, the visual network dynamically selects a portion of modules from that layer to process the input according to the binary gates generated by the routing network. We name the collection of all binary gates as a routing path. 
Therefore, the routing network is responsible for receiving the question features generated by the textual network and mapping them to a discrete routing path. To encourage the model to explore more possible routing paths, the model generates the routing path in a discrete and stochastic way by adopting Gumbel-Softmax tricks. We claim that our proposed method can address both two before-mentioned challenges to some extent. First, the visual and textual modalities can be fully fused at multiple semantic levels because the routing network controls every layer of the visual network. Second, the routing mechanism makes the model have the ability for visual reasoning by executing different portion of modules. Especially we claim that the routing path represents the process of compositional reasoning and acts similar to the aforementioned module layout.

To sum up, the contributions of this paper can be summarized as: (1) We for the first time successfully extend the line of work on routing models to the domain of VQA and multi-modal embedding, and the results are promising. (2) Our model can fuse the visual and textual modalities at multiple semantic levels, and is equipped with good reasoning ability. (3) Current modular network based VQA methods have great interpretability, but they also have the disadvantages of requiring expert knowledge or extra supervision information. Without having those disadvantages, we combine the routing mechanism with the modular network to obtain the modular routing network. The novel modular routing network can be applied to many off-the-shelf CNN models. (4) The proposed model has achieved state-of-the-art performances on the challenging CLEVR~\cite{johnson2017clevr} and CLEVR-Humans~\cite{johnson2017inferring} datasets.

\section{Related Work}
\subsection{Fusion Strategy of VQA}
Usually, the first step of most VQA methods is to extract high-level visual features from a modern CNN and textual features from an RNN separately. In order to combine the visual and textual modalities to produce the answer, many methods have been proposed to fuse the extracted visual and textual features.
Multimodal Low-rank Bilinear pooling (MLB)~\cite{Kim2016MultimodalRL} provides an efficient method to approximate the full bilinear pooling by forcing the weight matrix to be low-rank.  Multimodal Compact Bilinear pooling (MCB)~\cite{Fukui2016MultimodalCB} randomly projects the visual and textual features into a higher dimensional space, then processes them in Fast Fourier Transform space. Multimodal Tucker Fusion (MUTAN)~\cite{ben2017mutan} proposes a general fusion method based on Tucker decomposition, which covers MLB and MCB.
Yang \emph{et al.}~\cite{yang2016stacked} propose a multi-hop spatial attention to fuse the visual and textual features so that the image regions related to the question will be focused. Anderson \emph{et al.}~\cite{anderson2018bottom} propose a bottom-up model that combines the attention mechanism with object-level visual features, so that objects related to the question will be focused.
However, with only the high-level multi-modal features fused, these fusion methods do not fully utilized the multi-level interactions between two modalities.

Another line of work proposes to fuse two modalities by using the question to predict the parameters of the visual network. Due to the large number of parameters in the visual network, only a small portion of parameters can be learned from the question feature. For example, Gao \emph{et al.}~\cite{gao2018question} propose the Question-guided Hybrid Convolution (QGHC) based on group convolution, which consists of question-dependent kernels and question-independent kernels; the parameters of batch-norm~\cite{ioffe2015batch}  layers in MODERN~\cite{de2017modulating} are predicted by the question; FiLM~\cite{perez2018film} implements the condition through a general feature-wise transformation. Although this kind of methods can fuse the visual and textual modalities in a multi-level and fine-grained way, these methods still have limitations. If too many parameters are learned from the question, that will make the model difficult to train, but if only a few parameters are learned from the question, that will constrain the model's learning capacity. Another concern is about the flexibility. In particular, QGHC can only be applied in the ResNext~\cite{Xie2017AggregatedRT} architecture, MODERN can only be applied in CNN with the batch-norm layers.

Our proposed model can fuse the two modalities at multiple semantic levels, and more importantly, our model is very flexible. When we adjust the granularity of the module to filter level, MODERN and FiLM are related to our model, yet with fundamental differences. First, they modulate the visual network at a feature level, but we modulate the visual network at a module level. Second, we route the modules in a discrete and stochastic way, which means that our approach can explore more routing paths and is a step towards discrete reasoning.

\subsection{Visual Reasoning in VQA}
Judging whether a VQA model has reasoning ability is usually based on whether the model can well solve the CLEVR dataset, since the questions in CLEVR dataset are quite complicated that require reasoning to answer. There have been several prominent models that well solve the CLEVR dataset. Except for RN~\cite{santoro2017simple} that supports reasoning explicitly by utilizing pairwise comparisons, most methods are based on the idea of compositional reasoning, implicitly~\cite{perez2018film,hudson2018compositional} or explicitly~\cite{andreas2016neural,hu2017learning,johnson2017inferring,mascharka2018transparency,mao2019neuro}. The representative method of explicit compositional reasoning is Neural Module Networks (NMN)~\cite{andreas2016neural} which is dynamically instantiated from a collection of reusable modules based on the compositional structure of the question. Although the function of each module is learned from training, the question parser and the mapping rules from the parsing tree node to the module must be pre-defined. The performance of the NMN model heavily relies on the quality of the question parser chosen. Further, Hu \emph{et al.}~\cite{hu2017learning} proposes an End-to-End Module Network which predicts the module layout by an LSTM instead of an external question parser. Johnson \emph{et al.}~\cite{johnson2017inferring} proposes a model combining with both program generator and execute engine based on neural module network. TbD-net~\cite{mascharka2018transparency} combines neural modules and attention mechanism to achieve state-of-the-art accuracy on the CLEVR dataset. Different from above methods, NS-CL~\cite{mao2019neuro} can achieve state-of-the-art performance even without extra supervision information.

Although the NMN family has good interpretability, they usually require expert knowledge for designing the model and extra supervision information for training. Our method shares the concept of the module with the NMN family, however, the modules from our method are not dedicated-designed and no extra supervision information is required for training, we also combine the modular network with the routing mechanism.
Compared with the explicit reasoning models like FiLM~\cite{perez2018film} and MAC~\cite{hudson2018compositional}, our method uses a completely different routing mechanism to support reasoning.

\subsection{Routing Models}
The proposed modular routing network is related to conditional computation (CC)~\cite{bengio2013estimating,Bengio2015ConditionalCI} and mixture-of-experts (MoE)~\cite{jordan1994hierarchical,jacobs1991adaptive,eigen2013learning,Shazeer2017OutrageouslyLN}. CC refers to the dynamic execution of a part of the model based on the input. Bengio \emph{et al.}~\cite{bengio2013estimating} first propose the concept of CC and introduce four approaches to propagate gradients through stochastic neurons to support CC. Bengio \emph{et al.}~\cite{Bengio2015ConditionalCI} propose by using reinforcement learning as a tool to optimize CC, they also propose a regularization mechanism to encourage sparsity and diversity. MoE related research can be traced back to the 1990s~\cite{jordan1994hierarchical,jacobs1991adaptive}, and usually, an expert is a whole model. Eigen \emph{et al.}~\cite{eigen2013learning} propose a deep learning model that is made up of two MoEs, Shazeer \emph{et al.}~\cite{Shazeer2017OutrageouslyLN} introduce the sparsely-gated MoE by using a noisy gating method. 
Kirsch \emph{et al.}~\cite{Kirsch2018ModularNL} use generalized Viterbi EM to enable training without artificial regularization.
Routing model is equivalent to doing CC on MoE. Researchers have tried to apply routing models to multi-task learning~\cite{rosenbaum2018routing} and classification problems~\cite{Wu2018BlockDropDI,Veit2018ConvolutionalNW}. However, these methods are either adopting small-scale models, experimenting on toy-dataset, or achieving little improvement on the basis of static models. 

As Ramachandran \emph{et al.} said in their paper~\cite{ramachandran2018diversity}, although routing models are a promising direction of research, there must be successful applications of routing models where static models struggle. In this paper, we find that VQA (especially those who need strong reasoning ability) is one of the successful applications, the static models (\textit{i.e.,} standard VQA methods) perform poorly on the VQA datasets that require reasoning, but the routing models outperform them by a large margin.

\begin{figure*}[tb]
	\centering
	\includegraphics[width=1\textwidth]{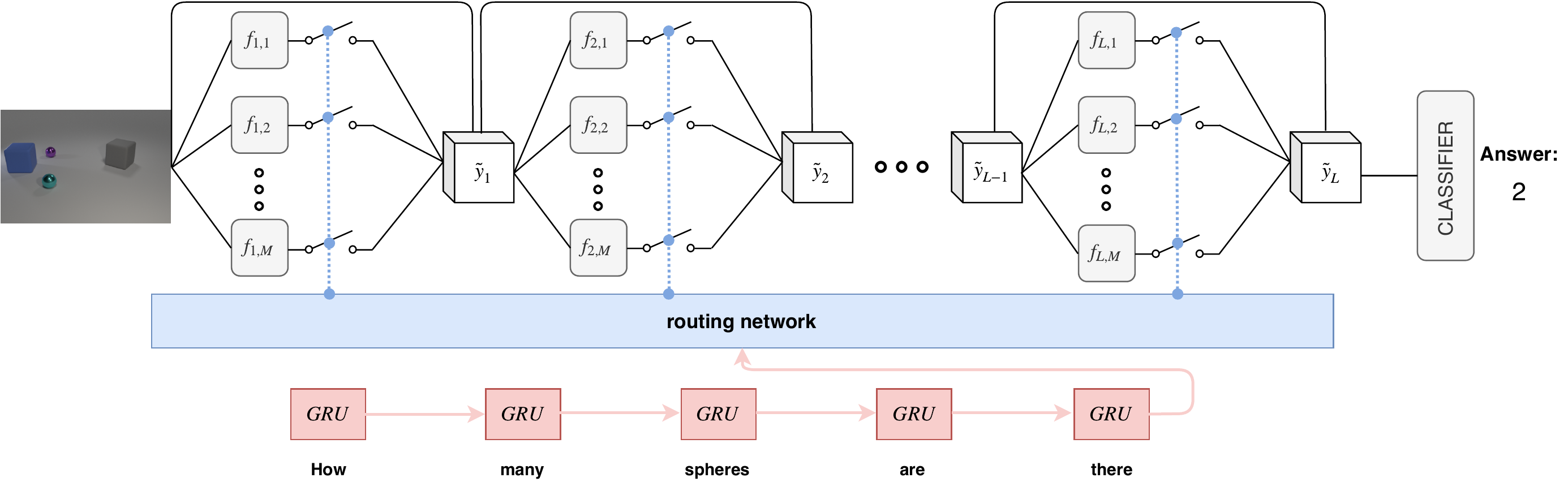}
	\caption{Architecture of the proposed model. The model consists three sub-networks: a) Visual Network: a composite of $L\times M$ modules which are represented as $f_{1..L,1..M}$, we can see that each module is followed by a ``switch'' which is controlled by the Routing Network, we call the states of the ``switches'' as a routing path. b) Textual Network: a GRU which takes a question and generates question features. c) Routing Network: which takes the question features sent from the textual network to generate a routing path.}
	\label{fig:arch}
\end{figure*}

\section{METHODOLOGY}

Our model contains three sub-networks, namely the \textit{visual network}, \textit{textual network} and \textit{routing network} as shown in Figure~\ref{fig:arch}. 
The textual network takes a question and generates question features; 
the routing network converts the question features to a specific routing path; 
the visual network takes the raw image and dynamically executes modules according to the routing path. 
Finally, the extracted image features are sent to the classifier to predict the answer.
The whole model is differentiable with respect to the model parameters and can be trained end-to-end.

\subsection{Textual Network}
Given the question $\matr{Q} = [\vect{w}_1, \vect{w}_2, \cdots, \vect{w}_n]$ where $\vect{w}_i$ is the one-hot representation of $i$th word and $n$ is the length of the question, we first use a lookup layer to embed $\matr{Q}$ into $\matr{E}_q = [\vect{e}_1, \vect{e}_2, \cdots, \vect{e}_n]$ where $\vect{e}_i \in \mathbb{R}^d$ , $d$ is the embedding size. Then we feed $\matr{E}_q$ into Gated Recurrent Units (GRU) and use the final hidden state of GRU as the question features:
 \begin{equation}
    \vect{q} = \text{GRU}(\matr{E}_q)
 \end{equation}
where $\vect{q} \in \mathbb{R}^h$ and $h$ is the hidden size of GRU units.

 \subsection{Visual Network}
 To better illustrate our approach, we will first introduce the generic Modular Architecture and Modular Routing Architecture. Modular Architecture defines the granularity and organization structure of the modules. Module Routing Architecture defines how the routing path controls over the Modular Architecture. Then we will introduce two forms of implementations of the module routing architecture:  \textbf{FRN}, which is based on ResNet; \textbf{BRN} (\textbf{B}ranch \textbf{R}outing \textbf{N}etwork), which is based on ResNext~\cite{Xie2017AggregatedRT}.
 
\subsubsection{Modular Architecture.}
We assume that a generic modular architecture is composed of $L$ modular layers, each modular layer is composed of $M$ modules\footnote{For convenience, we assume that the number of modules per modular layer is the same, but in fact they can be different.}, each module $f_{l,m}(x;\theta_{l,m})$  for $l \in \{ 1, \cdots, L \}$ and $m \in \{ 1, \cdots, M \}$ is a function module that takes the input $x$ and generates an output tensor, $\theta_{l,m}$ is the learnable parameter of $f_{l,m}$. All modules in the $l$th layer share the same input, and the output of the $l$th layer $y_l$ is:
\begin{equation}
    y_l = \phi([f_{l,1}(y_{l-1};\theta_{l,1}), \cdots, f_{l,M}(y_{l-1};\theta_{l,M})]) + y_{l-1}
 \end{equation}
where the composite function $\phi(.)$ can either be concatenation or summation.

\subsubsection{Modular Routing Architecture}
First, we denote the routing path as
 \begin{equation}
    \matr{P} = \text{RNET}(\vect{q})
 \end{equation}
where RNET is the routing network that generates a routing path $\matr{P} \in \{0,1\}^{L\times M}$ condtioned on the question features $\vect{q} \in \mathbb{R}^h$.  $\matr{P}_{l,m}$ for $l \in \{1, \cdots, L\}$ and $m \in \{ 1, \cdots, M \}$ is a binary gate that controls whether or not to execute the $m$th module of the $l$th layer. The output of the $l$th layer $\tilde{y}_l$ of modular routing architecture is now changed to:
\begin{equation}
	 \tilde{y}_l = \phi([\tilde{f}_{l,1}(\tilde{y}_{l-1};\theta_{l,1}), \cdots, \tilde{f}_{l,M}(\tilde{y}_{l-1};\theta_{l,M})]) + \tilde{y}_{l-1}\\
\end{equation}
where $\tilde{f}_{l,m}(\tilde{y}_{l-1};\theta_{l,m}) = \matr{P}_{l,m}\cdot f_{l,m}(\tilde{y}_{l-1};\theta_{l,m})$.
Due to the discrete nature of $\matr{P}$, the gradient backpropagation algorithm cannot be applied here, we will use some tricks to make the whole model differentiable and the details will be discussed in section~\ref{rn_sec}.

\subsubsection{FRN and BRN}
As we claimed before, with the introduced modular routing architecture, different models can be realized by changing the granularity of the module. Here we introduce two kinds of special cases that are easy to implement.\\
\noindent \textbf{FRN: }when the routed module is the filter in a convolutional layer, we call it FRN. Note that the FRN can be plugged into any modern CNN, we applied it to the current popular ResNet for convenience. More specifically, we route the last convolutional layer of each \textit{residual block}.\\
\noindent \textbf{BRN: }when the routed module is the branch of a multi-branch network, we call it BRN. The BRN can be plugged into any multi-branch convolutional networks. In this paper, we test the BRN under the ResNext model.

\subsection{Routing Network}
\label{rn_sec}
As we claimed before, RNET aims to compute the binary gates conditioned on the question features. First, we temporarily ignore the fact that the routing path is discrete for simplicity, All we need is a fully connected layer $\text{fc}: \mathbb{R}^h \rightarrow \mathbb{R}^{L\times M}$ to map the question features to a \textit{real-valued} routing path $\matr{\tilde{P}}=\text{fc}(\vect{q})$. Now take into account the nature of the discrete, a naive attempt would be thresholding $\matr{\tilde{P}}$ into $1$s and $0$s, but unfortunately, it is not differentiable and the backpropagation algorithm cannot be applied. Also note that the threshold function is deterministic, in order to explore more possible paths, the generation of routing path will better to be stochastic. Based on the above considerations, \textit{i.e.,} \textbf{discrete} and \textbf{stochastic}, we let the routing path $\matr{P}$ follows a $L\times M$-dimensional Bernoulli distribution whose parameter $\matr{s} \in [0,1]^{L\times M}$ is generated from the routing network. To optimize the whole model, we employ a reparameterization trick called \textit{Concrete Distribution}~\cite{maddison2016concrete} or \textit{Gumbel- Softmax}~\cite{jang2016categorical} in this paper. 

In order to elicit this method, we first review the Gumbel-Max trick~\cite{yellott1977relationship} which provides a way to sample $\vect{z}$ from a categorical distribution with class probability of $\pi_1, \pi_2, \cdots, \pi_n$ as follows:
\begin{equation}
    \vect{z} = \text{one-hot}(\argmax_{i}{[ \textnormal{g}_i + \log \pi_i ]}) \\
\end{equation}
where $\textnormal{g}_1,...,\textnormal{g}_n$ are i.i.d samples from Gumbel distribution \footnote{To sample from Gumbel distribution, first draw a sample from Uniform distribution $\textnormal{u} \sim \text{Uniform}(0,1)$, computing random variable $\textnormal{g}$ as $\textnormal{g}=-\log(-\log(\textnormal{u}))$, then $\textnormal{g} \sim \text{Gumbel}(0,1)$} and $P(\vect{z}_k=1) = \pi_k $. However the argmax operation is still not differentiable, so the softmax function with temperature $\tau$ is introduced here to approximate the argmax function:
\begin{equation}
	\vect{\tilde{z}} = \text{softmax}((\textnormal{g}+\log(\pi))/\tau)
\end{equation}
As $\tau \rightarrow 0$, the softmax function is smoothly approaching the argmax function.

In the above we have showed how to sample from categorical distribution $\text{Cat}(\pi_1, \pi_2, \cdots, \pi_n)$ using the Gumbel-Softmax trick, now we discuss the ``binary'' case of our problem, that means we need to sample from $\text{Cat}(\pi_1, \pi_2)$ where $\pi_2=1-\pi_1$. Note that this distribution is equivalent to Bernoulli distribution $\text{Bern}(\pi_1)$, we use $\rho$ to substitute $\pi_1$ to distinguish from the previous symbol definition, then we apply the Gumbel-Max trick again to sample Bernoulli variable $\textnormal{b}$ from  $\text{Bern}(\rho)$ as
\begin{align}
\textnormal{b} &= [\textnormal{g}_1+\log(\rho) > \textnormal{g}_2+\log(1-\rho)] \\
				&= [\textnormal{g}_1-\textnormal{g}_2+\log(\rho/(1-\rho))>0 ] \\
	\label{eq:logistic}			&= [ \textnormal{t}+\log(\rho/(1-\rho))>0 ]
\end{align}
where $\textnormal{t} \sim \text{Logistic}(0,1)$ and the bracketed notation [statement] stands for $1$ if statement is true, $0$ otherwise.
The derivation of Eq.(\ref{eq:logistic}) uses the fact that the difference of two Gumbels variables follows a Logistic distribution\footnote{To sample from Logistic distribution, computing random variable $\textnormal{t}$ as $\textnormal{t}=\log(\textnormal{u})-\log(1-\textnormal{u})$ where $\textnormal{u} \sim \text{Uniform}(0,1)$, then $\textnormal{t} \sim \text{Logistic}(0,1)$}, \textit{i.e.,} $\textnormal{g}_1-\textnormal{g}_2 \sim \text{Logistic}(0,1)$. Just as we use the softmax function with temperature to approximate the argmax function, here we use the sigmoid function with temperature to approximate the unit step function:
\begin{equation}
\tilde{\textnormal{b}} = \text{sigmoid}((\textnormal{t}+\log(\rho/(1-\rho)))/\tau)
\end{equation}
as $\tau \rightarrow 0$, the sigmoid function is smoothly approaching the unit step function. Readers who are interested in Gumbel-Softmax or Concrete Distribution are referred to~\cite{maddison2016concrete,jang2016categorical} for more details.



Based on previous discussion, we can convert the \textit{real-valued} $\matr{\tilde{P}}$ to binary value in a simple way. First, for each entry $\matr{\tilde{P}}_{l,m}$, we compute $\matr{\tilde{B}}_{l,m}$ as:
\begin{equation}
\matr{\tilde{B}}_{l,m} = \text{sigmoid}((\textnormal{t}_{l,m}+\matr{\tilde{P}}_{l,m})/\tau)
\end{equation}
where $\textnormal{t}_{1...L,1...M}$ are i.i.d samples from Logistic distribution and $\matr{\tilde{P}}_{l,m}=\log(\frac{\matr{s}_{l,m}}{1-\matr{s}_{l,m}})$. However $\matr{\tilde{B}}_{l,m}$ is still a continuous value as $\tau > 0$, here we use a Straight-Through (ST) method introduced in~\cite{bengio2013estimating} to convert the continuous $\matr{\tilde{B}}_{l,m}$ to discrete $\matr{P}_{l,m}$, that is, during forward process, we use a threshold of $0.5$ to thresholding $\matr{\tilde{B}}_{l,m}$ to $0$ and $1$, but during the backward process, the gradient is normally passed to $\matr{\tilde{B}}_{l,m}$ just as the thresholding function is an identity function. We provide the pseudocode of the routing algorithm in appendix~\ref{appendix:pseudocode}.

\subsection{QGMRN with Attention}
\label{sec:att}
Benefiting from the routing mechanism, the visual feature map $\tilde{y}_L$ extracted from the visual network already contains enough information and clues to answer the question, so we can directly send it to a global max pooling (GMP) layer and a simple MLP classifier to predict the answer. 

And we can also use an additional attention mechanism to further fuse the multi-modal features. The attention method consists of two stages: (1) We use the spatial self-attention mechanism to enhance the representation power and  the spatial reasoning ability of the model; (2) We use a simple spatial attention mechanism to convert the feature map to the final feature vector.

In the first stage, we first concatenate the question features $\vect{q}$ with the visual features $\tilde{y}_L$ to get the fused multi-modal features $u \in \mathbb{R}^{h\times w\times c}$ (broadcasting is used here) where $h$, $w$, and $c$ denote the height, width and channel of the feature map. Then we reshape it to $hw \times c$, and feed it to an off-the-shelf Transformer~\cite{vaswani2017attention} encoder layers for modeling the correlations among the spatial locations. After we get the output $\tilde{u} \in \mathbb{R}^{hw \times d_t}$ where $d_t$  is the dimension of the Transformer, we reshape it back to $h\times w \times d_t$. 

In the second stage, rather than using GMP as aggregator to convert the feature map $\tilde{u}$ to the feature vector, we find the weighted summation as aggregator is better. Specifically, we send $\tilde{u}$ to a $1\times1$ convolution layer followed by a spatial softmax layer to get the attention/weight map $a \in \mathbb{R}^{h\times w\times 1}$, the final features $f$ send to the MLP classifier is the weighted sum of $\tilde{u}$ at all spatial locations:
\begin{equation}
	f=\text{sum}(a\cdot\tilde{u}, \text{dim}=(0, 1))
\end{equation}
where $f\in \mathbb{R}^{d_t}$.

%
%

\subsection{Training Loss}
In order to avoid model collapse and prevent certain modules from being always executed or always not executed, some sparsity and variance regularizations~\cite{Bengio2015ConditionalCI,Veit2018ConvolutionalNW,Shazeer2017OutrageouslyLN} are introduced. In this paper, we do not require our model to be sparse. But we observe that adding an regularization technique called load-balancing loss $\mathcal{L}_{load}$, which was first introduced in~\cite{Shazeer2017OutrageouslyLN} to prevent model collapse, will make the model converge faster. 
We define the load of a module as the number of samples in one training batch activate that module. So the load of the $m$th module in the $l$th layer is calculated as: $\matr{A}_{l,m}= \sum_{i=1}^{N}\matr{P}_{l,m}^{(i)}$, where $\matr{P}_{l,m}^{(i)}$ is the gate of the $m$th module in the $l$th layer for the $i$th instance in the mini-batch, $N$ is the batch size. Further, we define $\mathcal{L}_{load_l} = \mathrm{CV}(\matr{A}_{l,:})^2$ as the square of the coefficient of variation (CV) for the module load at the $l$th layer, and $\mathcal{L}_{load} = \sum_{l=1}^{L} \mathcal{L}_{load_l}$ as the sum of $\mathcal{L}_{load_l}$ in all layers.

With the standard VQA loss $\mathcal{L}_{vqa}$, the full loss is:
\begin{equation}
	\mathcal{L} = \mathcal{L}_{vqa} + \lambda \mathcal{L}_{load}
\end{equation}
where $\lambda$ is the coefficient of the load-balancing loss.

\section{EXPERIMENTS}

\subsection{Datasets}
\begin{table*}[!ht]
\centering
\small
\begin{tabular}{l | c c c c c c | c}
    \toprule
    {Model}  &\begin{tabular}{@{}c@{}}CLEVR \\ Overall\end{tabular} & {Count} & {Exist} & \begin{tabular}{@{}c@{}}Compare \\ Numbers\end{tabular} & \begin{tabular}{@{}c@{}}Query \\ Attribute\end{tabular} & \begin{tabular}{@{}c@{}}Compare \\ Attribute\end{tabular} & \begin{tabular}{@{}c@{}}Humans \\ After FT\end{tabular}\\
    \midrule
    Human~\cite{johnson2017inferring}                  & 92.6 &86.7 &96.6 &86.5 &95.0 &96.0 &\textendash{}\\
    \midrule
    N2NMN*~\cite{hu2017learning}              &83.7 &68.5 &85.7 &84.9 &90.0 &88.7 &\textendash{}\\
    PG+EE (18K prog.)*~\cite{johnson2017inferring}      &95.4 &90.1 &97.3 &96.5 &97.4 &98.0 &66.6\\
    PG+EE (700K prog.)*~\cite{johnson2017inferring}    &96.9 &92.7 &97.1 &98.7 &98.1 &98.9 &\textendash{}\\
    TbD*~\cite{mascharka2018transparency} &99.1 &97.6 &99.2 &99.4 &99.5 &99.6 &\textendash{}\\
    \midrule
    LSTM~\cite{johnson2017inferring}                   &46.8 &41.7 &61.1 &69.8 &36.8 &51.8 &36.5\\
    CNN+LSTM~\cite{johnson2017inferring}               &52.3 &43.7 &65.2 &67.1 &49.3 &53.0 &43.2\\
    CNN+LSTM+SA~\cite{johnson2017inferring}             &76.6 &64.4 &82.7 &77.4 &82.6 &75.4 &57.6\\

    FiLM~\cite{perez2018film}     &97.6 &94.3 &99.3 &93.4 &99.3 &99.3 &75.9\\
    QGHC~\cite{gao2018question}  &86.3 &78.1 &91.7 &80.7 &89.4 &86.8 &\textendash{}\\
    Relation Network~\cite{santoro2017simple}  &95.5 &90.1 &97.8 & 93.6 &97.9 &97.1 &\textendash{}\\
    MAC~\cite{hudson2018compositional}  &\cellcolor[HTML]{C4C7FD}{98.9} &97.2 &99.5 &\cellcolor[HTML]{C4C7FD}{99.4} &99.3 &99.5 &\cellcolor[HTML]{EBECFF}{81.5}\\
    NS-CL~\cite{mao2019neuro}  &\cellcolor[HTML]{C4C7FD}{98.9} &\cellcolor[HTML]{C4C7FD}{98.2} &98.9 &\cellcolor[HTML]{EBECFF}{99.0} &99.3 &99.1 &\textendash{}\\
    \midrule
    BRN (c$=$8) (ours)  & 86.8 & 79.5 & 93.4 & 75.1 & 90.0 & 91.1 &\textendash{}\\
    BRN (c$=$16) (ours)  &94.7 &90.8 &98.1 &82.8 &97.3 &98.4 &\textendash{}\\
    BRN (c$=$32) (ours)   &97.9 &95.4 &98.9 &92.9 &99.4 &\cellcolor[HTML]{EBECFF}{99.7} &77.9\\
    BRN (c$=$32)+Attention (ours)  &\cellcolor[HTML]{EBECFF}{98.7} &97.4 &\cellcolor[HTML]{EBECFF}{99.6} &95.1 &\cellcolor[HTML]{EBECFF}{99.7} &\cellcolor[HTML]{EBECFF}{99.7} &79.3\\
    FRN (ours)  &98.2 &96.6 &\cellcolor[HTML]{C4C7FD}{99.9} &94.1 &98.7 &\cellcolor[HTML]{C4C7FD}{99.9} &79.9\\
    FRN+Attention (ours)  &\cellcolor[HTML]{C4C7FD}{98.9} &\cellcolor[HTML]{EBECFF}{97.7} &\cellcolor[HTML]{C4C7FD}{99.9} &94.5 &\cellcolor[HTML]{C4C7FD}{99.8} &\cellcolor[HTML]{C4C7FD}{99.9} &\cellcolor[HTML]{C4C7FD}{81.8}\\
    \bottomrule
\end{tabular}
\caption{Comparison of accuracy on CLEVR and CLEVR-Humans datasets with previous methods. (*) denotes use of extra supervisory information through program labels. And c is the abbreviation for cardinality, and represents the number of modules per modular layer. The highest and second highest accuracies are shown in different background colors. However, we did not highlight TbD since it uses the extra supervisory information.}
\label{tab:accuracy}
\end{table*}

The proposed method is evaluated on two datasets: (1) CLEVR dataset~\cite{johnson2017clevr} is proposed to study the ability of VQA systems to perform reasoning. Answering question about a CLEVR image requires various kinds of reasoning, which makes the standard VQA methods perform poorly on this dataset. The dataset contains 100K 3D-rendered images and about one million automatically-generated questions. Specifically, the question in the dataset can be divided into the following five types: \texttt{Count}: ask the number of certain objects; \texttt{Exist}: ask whether a certain object is present; \texttt{Compare Numbers}: ask which of two object sets is larger; \texttt{Query Attribute}: query a attribute of particular object; \texttt{Compare Attribute}: ask whether two particular objects have same value on some attribute. (2) CLEVR-Humans dataset~\cite{johnson2017inferring} contains human-posed questions on CLEVR images, which makes the dataset more complex and realistic.
We train all of our models with the official training set and test the models on the official validation set, and we train our models from raw pixels.

\subsection{Implementation details}
\noindent \textbf{Configuration of BRN and FRN:} BRN has $8$ modular routing layers, and each layer has $8/16/32$ modules. In the terminology of the ResNext paper, the cardinality is $8/16/32$ and the depth is $26=3\times8+2$. It is notable that we replace the $3$rd batch normalization (bn) of the bottleneck block with group normalization (gn). Although most studies have shown that gn does not perform as well as bn when the batch size is large enough, and it is rarely seen that gn is applied to the ResNext model. But for BRN, the combination of gn and ResNext will greatly improve the performance of the model. We regard this as an interesting discovery, and we provide a possible explanation in the ablation study. FRN is based on ResNet34, that is, each modular layer is a \textit{basicblock} and has $16$ modular layers in total, the number of modules for each layer depends on the second convolutional layer's filter number.

\vspace{0.5mm}
\noindent \textbf{Visual Network:} all images are resized to $480\times 360$ and we found that whether the image size is $480\times 360$ or $224\times 224$ has little effect on the performance of BRN, so BRN takes images of $224\times 224$ as input. We also found that sending the question features together with the visual features into the classifier will help the BRN training more stable. In order to achieve the best performance, we concatenate $\tilde{y}_L$ with two coordinate feature maps indicating relative $x$ and $y$ spatial position.

\vspace{0.5mm}
\noindent \textbf{Textual Network:} the word embedding size is set to $200$, the GRU hidden size is set to $512$, we observe that the hidden size set to $512$ or $1024$ has little effect on the final accuracy. The parameters of the GRU and word embedding layer are initialized with orthogonal initialization and uniform initialization respectively.

\vspace{0.5mm}
\noindent \textbf{Routing Network:} although the temperature $\tau$ can be annealed to a small value during training, we find that just keeping it a constant value $1.0$ can get decent accuracy. Note that we need the generation of routing path to be stochastic to exploit more possible routing paths during training, but during test phrase, we want the generation of routing path to be deterministic, so just use the sigmoid function with a threshold $0.5$ to convert the $\tilde{P}$ to $P$. We also initialize the parameters of the routing network so that the probability of each module being executed at the beginning is $0.7$, that is, the bias of the fc layer of routing network is initialized to $\log(\frac{0.7}{1-0.7})$.

\vspace{0.5mm}
\noindent \textbf{Training:} all the models are trained with ADAM\cite{kingma2014adam} optimizer, betas are set to $(0.9, 0.999)$, batchsize is set to 64, learning rate is set to $3e-4$. We observe that applying a warmup scheme\cite{goyal2017accurate} can help the FRN to achieve better performance, \textit{i.e.,} we start training our model with a small learning rate 3e-6, and slowly increase the learning rate until it reaches 3e-4, then use $3e-4$ to train the model until the end.

\subsection{Results on CLEVR dataset}

The results of all the compared methods on CLEVR are shown in Table~\ref{tab:accuracy}, as we claimed before, the standard VQA methods like CNN+LSTM, CNN+LSTM+MCB, CNN+LSTM+SA perform poorly on this challenging dataset. Compared with the explicit reasoning methods N2NMN, PG+EE, TbD, and NS-CL, which are also based on modular networks, most of our proposed models outperform N2NMN and PG+EE, and our best-performed model is as good as NS-CL. Please note that all the four explicit methods require expert knowledge to analyze subquestion set and design corresponding modules for solving them, and all methods except NS-CL require extra supervisory information. So although it is slightly unfair to compare our models with these methods, our models still achieve competitive performance. We attribute such results to the combination of routing mechanism and modular networks. Compared with question guided methods like FiLM and QGHC, our question guided modular routing networks still outperform these methods, which validates the effectiveness of routing mechanism, especially that QGHC and BRN are both based on ResNext, but BRN outperforms QGHC by a large margin. Compared with the implicit reasoning methods like FiLM and MAC, and the prominent Relation Network, our best-performed model still outperforms most of these methods and achieves the same level of performance as MAC.

By comparing the three BRN models with different branch numbers, we can basically conclude that increasing the number of modules per layer will improve the performance accordingly. This makes sense, as the diversity of routing path can be expressed as $2^{L*M}$ (the routing path consists of $L*M$ switches, each of which has two states), if we keep the number of module layers $L$ consistent and increase the number of modules per layer $M$, the diversity of routing path will also increase. Since the diversity of routing path largely reflects the model's representation power, so increasing the number of modules per layer can enhance the performance of the model finally. We hope this discovery can provide some help for future research on routing models. By comparing the model performance with or without attention component we propose in Section~\ref{sec:att} , we can find the attention component can improve the overall accuracy by about 0.7\%.

Finally, on \texttt{Exist}, \texttt{Query Attribute} and \texttt{Compare Attribute} question types, our best-performed model has reached nearly 100\% accuracy, higher than all other methods. But on \texttt{Compare Numbers} question type, our models do not perform well. However previous study~\cite{perez2018film} has pointed out that on \texttt{Compare Numbers} question type, model trained from feature map (extracted from a pre-trained model) will perform better than the model trained from raw pixel.

\subsection{Results on CLEVR-Humans}
To further validate the reasoning ability and generalization ability of our model, we next provide the results on CLEVR-Humans. To make a fair comparison with other methods, the reported result is fine-tuned from previous best-performed BRN and FRN models on CLEVR dataset. Also, pre-trained word embeddings are not used.

As shown in the last column of Table~\ref{tab:accuracy}, our best-performed model outperforms all compared methods. The performance difference is more significant compared to CLEVR which validates that our model has better robustness and strong reasoning ability.

\subsection{Ablation study of BRN}
We conduct ablation studies of BRN on CLEVR dataset to analyze the influence of each decision to the model performance.
 
\vspace{0.5mm}
\noindent \textbf{Should we send question features to the classifier with image features?} As shown in Table~\ref{tab:ablation}, when the number of modules per modular layers is set to $16$, sending question features together with image features to the classifier will provide 3.3\% improvement over sending only image features to the classifier. However, we did not find a similar improvement in FRN, perhaps because the performance of FRN itself is high enough.

\vspace{0.5mm}
\noindent \textbf{Should we replace the 3rd normalization layer of the ResNext block from bn to gn?} First of all, why use gn instead of bn when the batchsize is already large enough? Since bn keeps $C$ pairs of mean and variance for each channel within a mini-batch. And a generally accepted rule is that the model with bn does not work well when the batchsize is too small. One possible reason is that the smaller batchsize will cause the statistics (mean and var) unstable, and change the distribution of feature map all the time. In the scenario of dynamic routing, bn's statistics will also be unstable. To validate our analysis, we replace the $3$rd normalization layer of the bottleneck block from bn to gn. The results can be found in Table~\ref{tab:ablation}, 
we can see the improvements are very significant after the replacement. As for why FRN can achieve good results without the replacement, we suspect that it may be due to the higher module execution ratio of FRN (please refer to section~\ref{execution}).

\vspace{0.5mm}
\noindent \textbf{How to select the number of modules per layer?} In the previous section, we have shown that for BRN with gn, increasing the number of modules per layer will improve the performance accordingly. In Table~\ref{tab:ablation}, we also show that the above rule apply to BRN without gn (\textit{i.e.,} without replacing the 3 normalization layers in the ResNext block) too.

\vspace{0.5mm}
\noindent \textbf{Will deepening or widening ResNext bring performance improvements?} The default depth and width of BRN's ResNext can be denoted as [2,2,2,2] and [64,128,256,256]. The $i$-th values of these two arrays represent the depth (\textit{i.e.,} the number of bottleneck blocks) and width (i.e., channel number) of the $i$-th stage, respectively. The $4$ stages are divided according to the downsampling operation. Based on this default configuration, we experiment with deepened (depth from [2,2,2,2] to [2,4,6,3]) or widened (width from [64,128,256,256] to [64,128,256,512]) models respectively. From Table~\ref{tab:ablation} we can find that deepening BRN will improve the accuracy by 0.3\%, however widening BRN will result in a 0.3\% accuracy drop. Therefore, the overall difference between the three models is not significant, but whether it is deepened or widened will increase the running time and slow down the convergence speed, so we finally use the default width and depth.

\setlength{\tabcolsep}{1.6pt}
\definecolor{Gray}{gray}{0.9}
\begin{table}[th]
\small
\centering
\begin{tabular}{l | c  c c c c c}
    \toprule
    {Model}  &{Overall} & {Count} & {Exist} & \begin{tabular}{@{}c@{}}Cmp. \\ Num.\end{tabular} & \begin{tabular}{@{}c@{}}Query \\ Attr.\end{tabular} & \begin{tabular}{@{}c@{}}Cmp. \\ Attr.\end{tabular}\\
    \midrule
    \multicolumn{7}{l}{\textbf{Ablation experiment on whether to feed $\mathbf{q}$ to the classifier}}\\
    \midrule
     \rowcolor{Gray} c=16, w/o q  &89.7 &84.7 &95.9 &81.9 &90.2 &94.6 \\
     \rowcolor{Gray} c=16, w/ q  &93.0 &90.8 &98.6 &83.6 &91.7 &98.9 \\
     
     \midrule
     \multicolumn{7}{l}{\textbf{Ablation experiment on whether to replace bn with gn}}\\
     \midrule
     c=8, w/ q &75.1 &66.8 &84.7 &75.8 &80.3 &68.4 \\
     c=8, w/ q, gn & 86.8 & 79.5 & 93.4 & 75.1 & 90.0 & 91.1  \\
     \rowcolor{Gray}c=16, w/ q &93.0 &90.8 &98.6 &83.6 &91.7 &98.9 \\
     \rowcolor{Gray} c=16, w/ q, gn & 94.7  & 90.8 & 98.1 & 82.8 & 97.3 & 98.4 \\
     c=32, w/ q &94.0 &93.0 &99.6 &85.9 &91.9 &99.4 \\
     c=32, w/ q, gn & 97.9 & 95.4 & 98.9 & 92.9 & 99.4 & 99.7 \\

    \midrule
    \multicolumn{7}{l}{\textbf{Ablation experiment on the number of modules per-layer}}\\
    \midrule
     \rowcolor{Gray} c=8, w/ q &75.1 &66.8 &84.7 &75.8 &80.3 &68.4 \\
     \rowcolor{Gray} c=16, w/ q &93.0 &90.8 &98.6 &83.6 &91.7 &98.9 \\
     \rowcolor{Gray} c=32, w/ q &94.0 &93.0 &99.6 &85.9 &91.9 &99.4 \\
    
      c=8, w/ q, gn & 86.8 & 79.5 & 93.4 & 75.1 & 90.0 & 91.1  \\
      c=16, w/ q, gn & 94.7  & 90.8 & 98.1 & 82.8 & 97.3 & 98.4 \\
      c=32, w/ q, gn & 97.9 & 95.4 & 98.9 & 92.9 & 99.4 & 99.7 \\

      \midrule
      \multicolumn{7}{l}{\textbf{Ablation experiment on increasing the depth and width}}\\
      \midrule
      \rowcolor{Gray} c=32, w/ q &94.0 &93.0 &99.6 &85.9 &91.9 &99.4 \\
      \rowcolor{Gray} c=32, w/ q, deeper &94.3 &94.1 &99.6 &85.4 &92.1 &99.5 \\
      \rowcolor{Gray} c=32, w/ q, wider & 93.7 & 91.7 & 99.0 & 83.8 & 93.0 & 99.1 \\
      

    \bottomrule
\end{tabular}
\caption{Ablation studies of BRN on CLEVR dataset. The notation c is the abbreviation for cardinality, and represents the number of modules per modular layer; w/ q represents sending both image features and question features to the classifier; w/o q represents only sending the image features to the classifier; gn denotes using group normalization instead of batch normalization; deeper means more bottleneck blocks and wider means more channel number.}
\label{tab:ablation}
\end{table}


\begin{figure*}[!ht]
\centering
\begin{minipage}[b]{0.6\textwidth}
	\centering
	\includegraphics[width=1.0\textwidth]{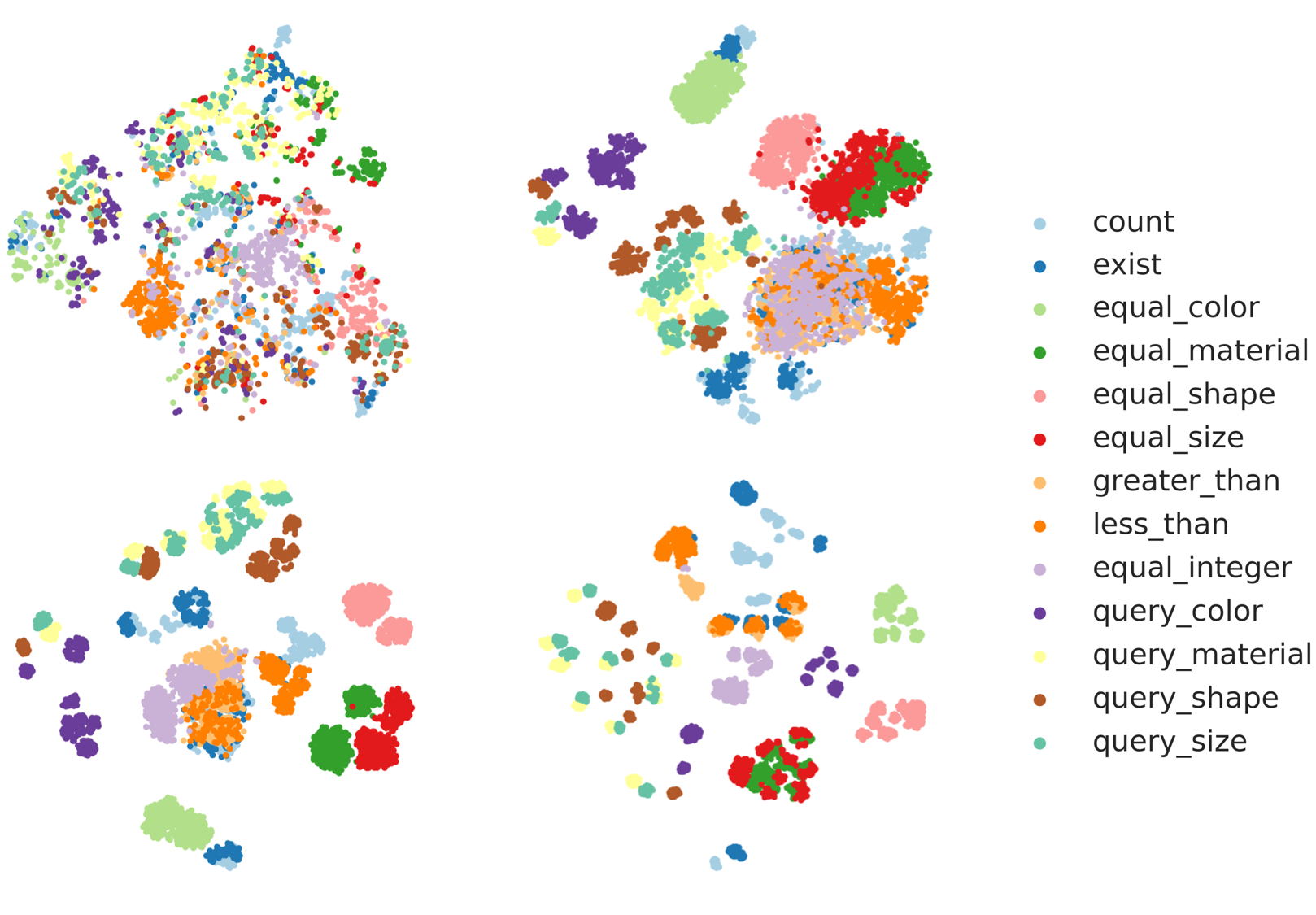}
	\caption{The t-SNE visualization of routing paths of stage1, stage2, stage3 and stage4 are represented at top-left, top-right, bottom-left and bottom-right sub-figures respectively. Low-level question subtypes are clustered in the earlier stages and high-level question subtypes are clustered in the later stages. Best seen on the computer, in colour and zoomed in.} \label{fig:tsne}
\end{minipage}
\hfill
\begin{minipage}[b]{0.32\textwidth}
	\centering
	\subfloat[FRN\label{fig:frn_ratio}]{\includegraphics[width=1.0\textwidth]{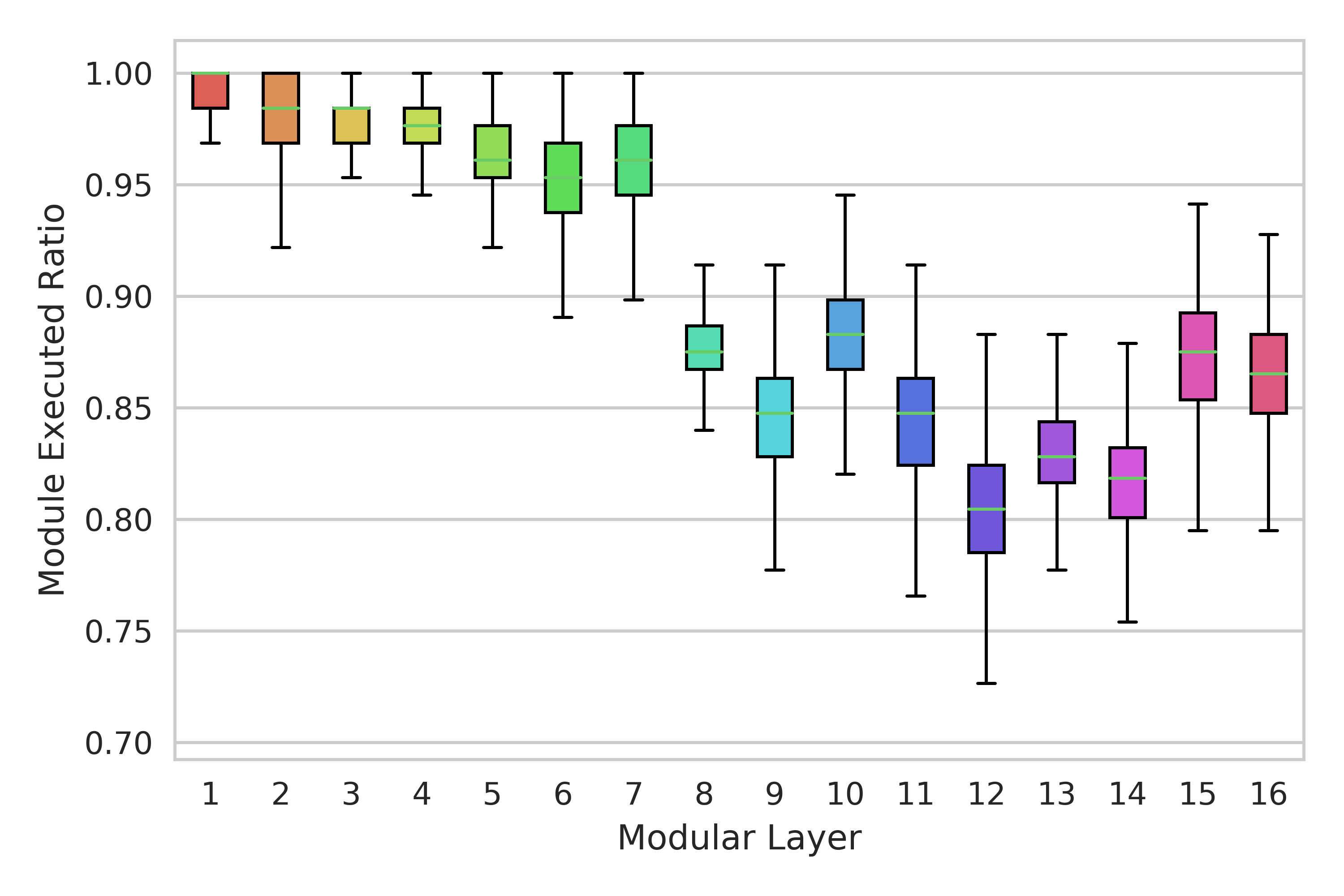}}\hfill
	\subfloat[BRN(c=32)\label{fig:brn_ratio}]{\includegraphics[width=1.0\textwidth]{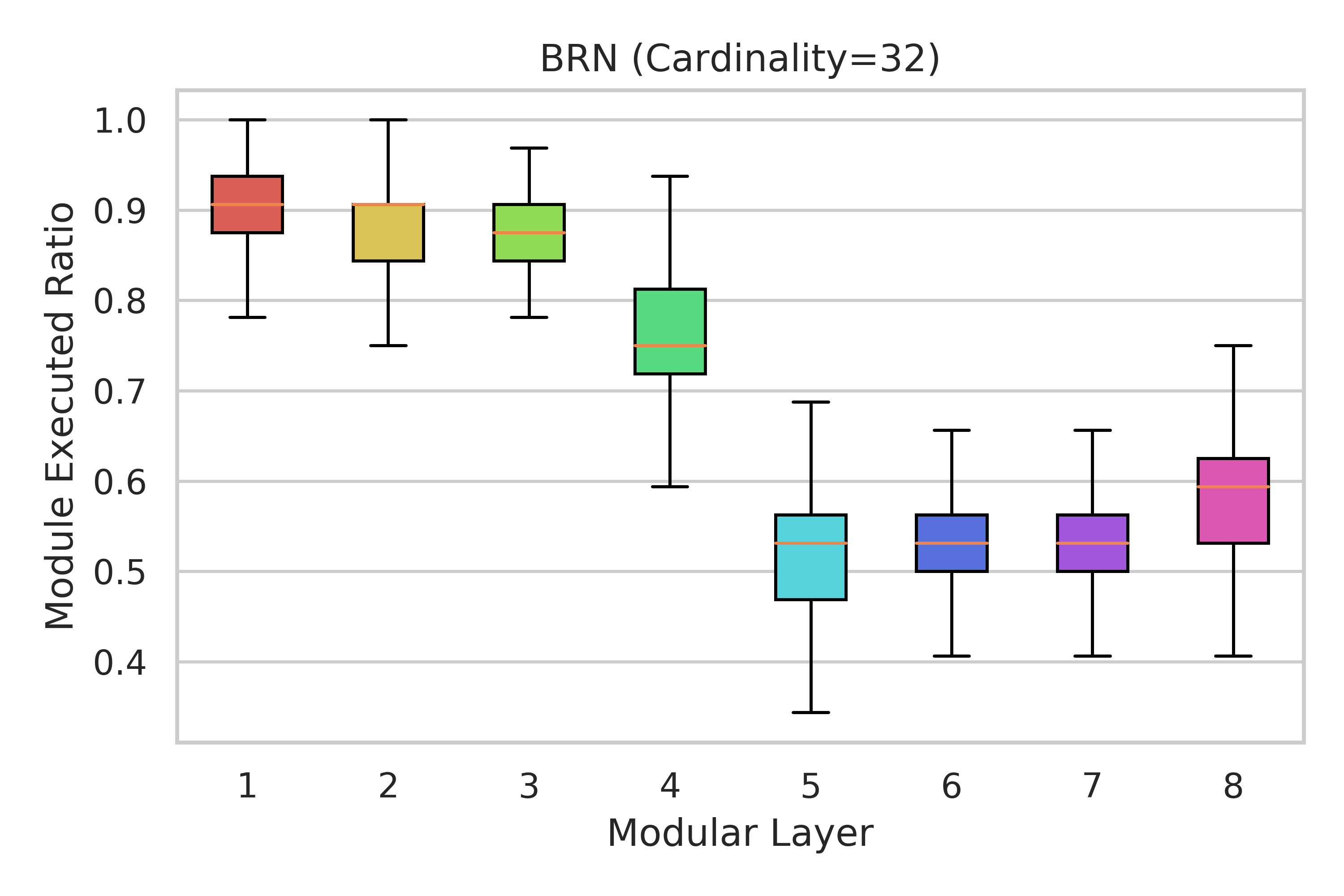}}\hfill
	\caption{The module execution ratio in each layer. The overall execution ratio of FRN is above BRN.} \label{fig:ratio}
\end{minipage}
\end{figure*}

\begin{figure}[tb]
	\centering
	\includegraphics[width=0.47\textwidth]{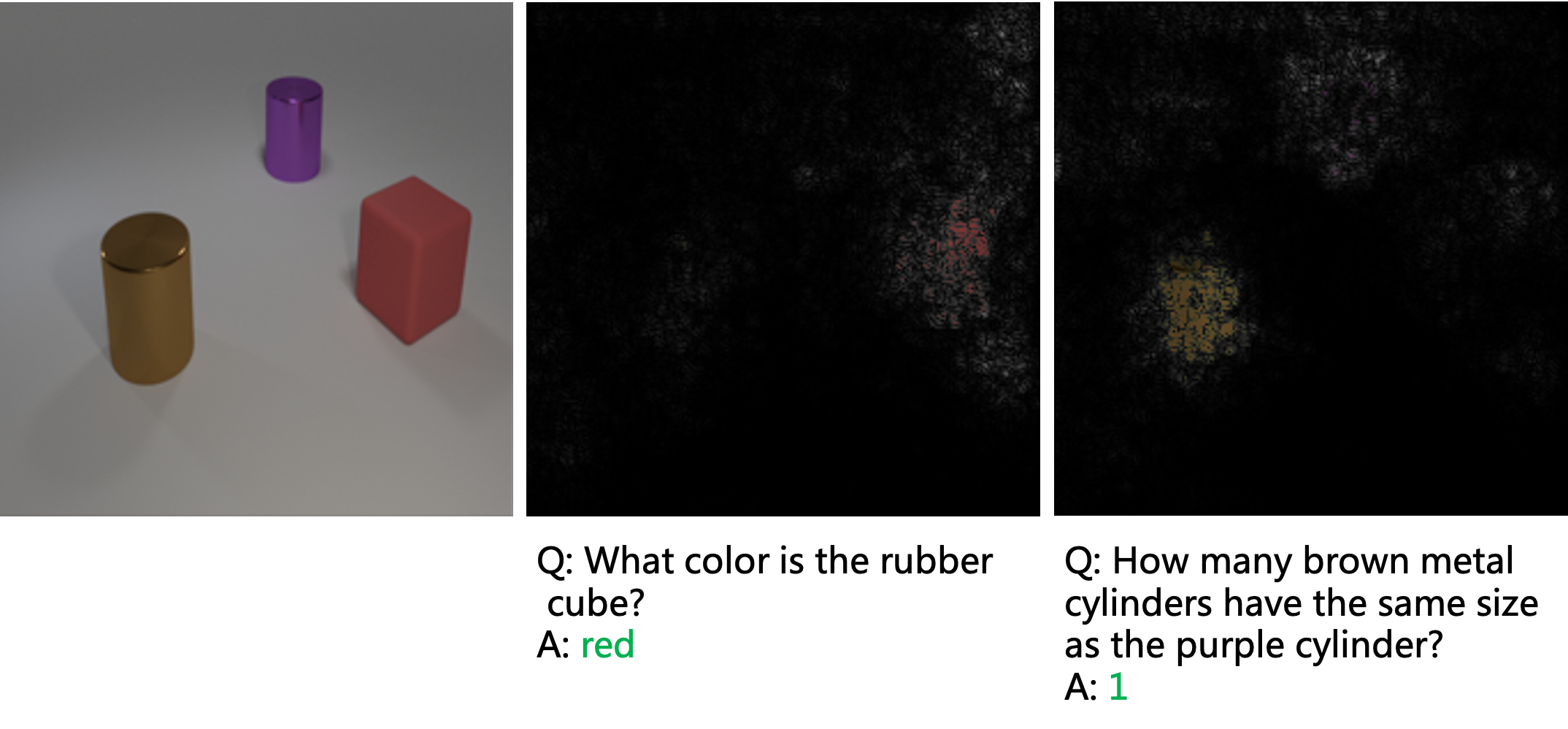}
	\caption{The saliency map indicates the key pixels for reasoning with respect to different questions.}
	\label{fig:saliency_example}
\end{figure}

\subsection{Visualization}

\noindent \textbf{T-SNE visualization of routing paths}\\
To investigate what the routing network learns, we use t-SNE~\cite{maaten2008visualizing} to visualize the routing paths in 2D embedding. Specifically, we divide the routing path into 4 stages according to the downsampling in the visual network, then flatten the routing path of each stage into a vector, and finally project it to the 2D space through t-SNE. For efficiency, we did not visualize the entire CLEVR dataset, but randomly select 500 instances on the validation set for each officially provided question subtype (\textit{i.e.,} subtypes subdivided from the previous 5 types). The visualization and detailed 13 question subtypes can be seen in Figure~\ref{fig:tsne}. Each point represents an instance, and points of the same question subtype are labeled with the same color. From the figure, we can discover some interesting phenomenons, we list some of them as follows: a) as the stage increases, data points with different question types can be discriminated better. b) The visualization of stage1 may be confusing at first glance, but note that data points with question subtype of \texttt{query\_color} and \texttt{equal\_color} are clustered together and data points with question subtype of \texttt{query\_material} and \texttt{equal\_material} are clustered together, this makes sense because the first few layers of CNN are responsible for detecting features about colors, textures, and edges~\cite{olah2017feature}. c) data points with the question subtype belonging to the same question type (\textit{e.g.,} \texttt{greater\_than}, \texttt{less\_than} and \texttt{equal\_integer} belong to the \texttt{Compare Numbers} type) are tend to cluster together. d) data points with the question type of \texttt{count} and \texttt{exist} are clustered together, this makes sense as \texttt{exist} is a special case of \texttt{count}.

\vspace{0.5mm}
\noindent \textbf{How many modules are executed for each layer}\\ \label{execution}
To investigate how many modules are executed in each layer, we provide the box plot (Figure~\ref{fig:ratio}) of the module execution ratio (\textit{i.e.,} the number of executed modules divided by the total number of the modules) for each layer of FRN and BRN(c=32) on the validation set. First of all, it can be observed that both FRN (refer to Figure~\ref{fig:frn_ratio}) and BRN (refer to Figure~\ref{fig:brn_ratio}) follow a rule: the execution ratio becomes lower as layers going deeper, this makes sense as the 
features extracted at early layers are generic, and the features extracted at higher layers are more related to the VQA task. Next we can observe that the overall execution ratio of the FRN model is not low, however please note that with the reduction of filters, current deep learning frameworks cannot reduce the running time accordingly, so we did not pay much attention to it. Finally, note that even if we do not add any sparsity constraints, BRN still has a considerable reduction of about \textbf{43\%}. We believe that this number can be increased with the addition of sparsity constraints. And with the further support of the deep learning framework, the routing model will have more useful applications in the future.

\vspace{0.5mm}
\noindent \textbf{Saliency map}\\
To investigate which pixels affect the prediction most, we visualize the saliency map of the same image to different questions. Concretely, we calculate the gradient of each pixel with respect to the max prediction score. Figure~\ref{fig:saliency_example} quantitatively shows that QGMRN can identify the key pixels for prediction. More examples are available in appendix~\ref{appendix:more_example}.

\section{Conclusion}
In this paper, we propose the Question Guided Modular Routing Networks for VQA which can fuse the visual and textual modalities at multiple semantic levels and learn to reason by routing between the generic modules, different interesting variant models can be generated by changing the granularity of the module. In the experiments, we show that our models have achieved the state-of-the-art performance. From the perspective of routing models, we find a suitable application for them where static models struggle, we also successfully applied them to large models and large-scale datasets. In particular, we provide two interesting findings: 1. The performance of the model is positively correlated with the number of modules per layer; 2. When the execution ratio is low, using gn will be better than bn. We believe that routing models will play an important role in future multimodal fusion and embedding.

%

%
\bibliographystyle{aaai}
\bibliography{aaai.bib}

%
\appendix

\section{Saliency map examples}
\label{appendix:more_example}
\begin{figure*}[htp]
    \centering
	\includegraphics[width=0.65\textwidth]{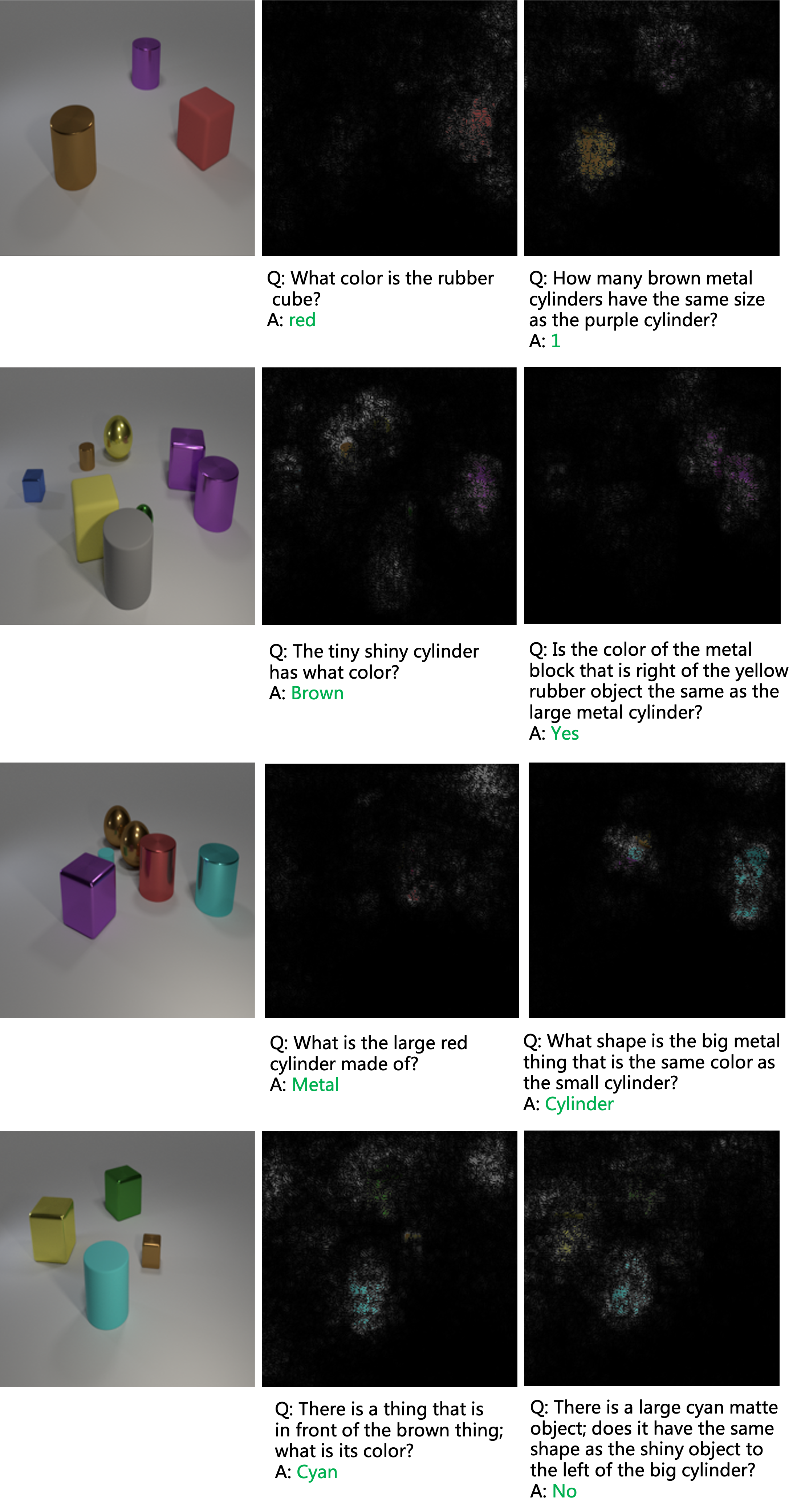}
	\caption{More saliency map examples with respect to different questions.}
	\label{fig:sal}
\end{figure*}
We provide several saliency maps to illustrate the key pixels conditioned on the question in Figure~\ref{fig:sal}. 

\section{Pseudocode of routing algorithm}
\label{appendix:pseudocode}
\begin{algorithm}
	\caption{Pseudocode of routing algorithm} 
	\label{alg:alg1}
	\begin{algorithmic}[1]
		\REQUIRE question feature $\vect{q}$; shape of routing path $L$ and $M$; temperature $\tau$
		\ENSURE routing path $\matr{P}$
		
		/* Sample from Logistic distribution */
		
		\STATE First Sample a random tensor from Uniform distribution $\textnormal{u} \sim \text{Uniform}([0, 1]^{L\times M})$, where $\textnormal{u} \in \mathbb{R}^{L\times M}$
		
		\STATE Then compute $\textnormal{t} \leftarrow \log(\textnormal{u}) - \log(1 - \textnormal{u})$, and $\textnormal{t} \sim \text{Logistic}([0, 1]^{L\times M})$
		
		/* Compute routing path */
		\STATE First use a fc layer to get the real-valued routing path $\tilde{\matr{P}} \leftarrow \text{FC}(\vect{q})$
		
		\STATE Then use Gumbel-Sigmoid trick to compute random tensor $\tilde{\matr{B}} \leftarrow \text{Sigmoid}(\frac{\textnormal{t} + \tilde{\matr{P}}}{\tau} )$
		\STATE Threshold $\tilde{\matr{B}}$ to get the $\tilde{\matr{B}}_{\text{binary}} \leftarrow \tilde{\matr{B}} > 0.5$
		\STATE Finally use Straight-Through (ST) trick~\cite{bengio2013estimating} to compute discrete routing path $\matr{P} \leftarrow \tilde{\matr{B}}_{\text{binary}} - \text{detach}(\tilde{\matr{B}}) + \tilde{\matr{B}}$
	\end{algorithmic}
\end{algorithm}

\end{document}